\documentclass[sigconf]{acmart}
\begin{document}

% Copyright
% \setcopyright{acmcopyright}
%\setcopyright{acmlicensed}
%\setcopyright{rightsretained}
%\setcopyright{usgov}
%\setcopyright{usgovmixed}
%\setcopyright{cagov}
%\setcopyright{cagovmixed}

\copyrightyear{2017} 
\acmYear{2017} 
\setcopyright{acmcopyright}
\acmConference{GeoRich'17 }{May 14, 2017}{Chicago, IL, USA}\acmPrice{15.00}\acmDOI{http://dx.doi.org/10.1145/3080546.3080547}
\acmISBN{978-1-4503-5047-1/17/05}
% DOI
% \doi{10.475/123_4}

% ISBN
% \isbn{123-4567-24-567/08/06}

%Conference
% \conferenceinfo{PLDI '13}{June 16--19, 2013, Seattle, WA, USA}

% \acmPrice{\$15.00}

%
% --- Author Metadata here ---
% \conferenceinfo{WWW}{'17 Perth, Australia}
%\CopyrightYear{2007} % Allows default copyright year (20XX) to be over-ridden - IF NEED BE.
%\crdata{0-12345-67-8/90/01}  % Allows default copyright data (0-89791-88-6/97/05) to be over-ridden - IF NEED BE.
% --- End of Author Metadata ---

\title{Using Contexts and Constraints for Improved Geotagging of Human Trafficking Webpages}

\author{Rahul Kapoor}
\affiliation{%
  \institution{Information Sciences Institute}
  \streetaddress{USC Viterbi School of Engineering}
  \city{Marina Del Rey} 
  \state{CA} 
  \postcode{90292}
}
\email{rahulkap@isi.edu}

\author{Mayank Kejriwal}
\affiliation{%
  \institution{Information Sciences Institute}
  \streetaddress{USC Viterbi School of Engineering}
  \city{Marina Del Rey} 
  \state{CA} 
  \postcode{90292}
}
\email{kejriwal@isi.edu}

\author{Pedro Szekely}
\affiliation{%
  \institution{Information Sciences Institute}
  \streetaddress{USC Viterbi School of Engineering}
  \city{Marina Del Rey} 
  \state{CA} 
  \postcode{90292}
}
\email{pszekely@isi.edu}
% \alignauthor
% Rahul Kapoor\\
%        \affaddr{Information Sciences Institute}\\
%        \affaddr{USC Viterbi School of Engineering}\\
%        \email{rahulkap@isi.edu}
% \alignauthor
% Mayank Kejriwal\\
%        \affaddr{Information Sciences Institute}\\
%        \affaddr{USC Viterbi School of Engineering}\\
%        \email{kejriwal@isi.edu}
% % 2nd. author
% \alignauthor
% Pedro Szekely\\
%        \affaddr{Information Sciences Institute}\\
%        \affaddr{USC Viterbi School of Engineering}\\
%        \email{pszekely@isi.edu}
% }
% There's nothing stopping you putting the seventh, eighth, etc.
% author on the opening page (as the 'third row') but we ask,
% for aesthetic reasons that you place these 'additional authors'
% in the \additional authors block, viz.

% Just remember to make sure that the TOTAL number of authors
% is the number that will appear on the first page PLUS the
% number that will appear in the \additionalauthors section.

\begin{abstract}
Extracting geographical tags from webpages is a well-motiva-ted application in many domains. In illicit domains with unusual language models, like human trafficking, extracting geotags with both high precision and recall is a challenging problem. In this paper, we describe a geotag extraction framework in which context, constraints and the openly available Geonames knowledge base work in tandem in an Integer Linear Programming (ILP) model to achieve good performance. In preliminary empirical investigations, the framework improves precision by 28.57\% and F-measure by 36.9\% on a difficult human trafficking geotagging task compared to a machine learning-based baseline. The method is already being integrated into an existing knowledge base construction system widely used by US law enforcement agencies to combat human trafficking.
\end{abstract}

%
% The code below should be generated by the tool at
% http://dl.acm.org/ccs.cfm
% Please copy and paste the code instead of the example below. 
%
% \begin{CCSXML}
% <ccs2012>
% <concept>
% <concept_id>10002951.10003260.10003277.10003279</concept_id>
% <concept_desc>Information systems~Data extraction and integration</concept_desc>
% <concept_significance>500</concept_significance>
% </concept>
% <concept>
% <concept_id>10002951.10003260.10003277.10003278</concept_id>
% <concept_desc>Information systems~Site wrapping</concept_desc>
% <concept_significance>300</concept_significance>
% </concept>
% </ccs2012>
% \end{CCSXML}

% \ccsdesc[500]{Information systems~Data extraction and integration}
% \ccsdesc[300]{Information systems~Site wrapping}

%
% End generated code
%

%
%  Use this command to print the description
%
% \printccsdesc

% We no longer use \terms command
%\terms{Theory}

\maketitle
%Reviewer Concerns (not in order): (1) give at least some examples for the reader to better understand the specific challenge. (2) baselines are weak (3) related work

\section{Introduction}\label{introduction}
The ubiquity of the Web has also had the unfortunate consequence of lowering the barrier of entry for players engaging in \emph{illicit} activities. One such activity is human trafficking. Although precise numbers for human trafficking Web advertising activity are not known, they are very high, possibly in the tens of millions of (not necessarily unique) advertisements posted on the Web \cite{ben}.

Recent advances in information extraction and knowledge base construction technology, especially using techniques like deep neural networks and word embeddings \cite{deepdive}, \cite{multitask}, gives investigators (such as law enforcement and intelligence agencies) the valuable opportunity to turn the Web against illicit players. Exploiting this opportunity for the human trafficking domain involves solving some specific challenges \cite{kejriwal2017information}, \cite{dig}.

First, human trafficking advertisements deliberately \emph{obfuscate} key pieces of information like names and phone numbers to avoid automated search, indexing and discovery. Second, like Twitter and social media, the \emph{language model} in human trafficking is non-traditional, using words, phrases and slang that impair performance of traditional extractors. As a representative example, consider the sentences \emph{`Hey gentleman im neWYOrk and i'm looking for generous...'} and \emph{`AVAILABLE NOW! ?? - (4 two 4) six 5 two - 0 9 three 1 - 21'}. In the first instance, the correct extraction for a \emph{Name} attribute is \emph{neWYOrk}, while in the second instance, the correct extraction for an \emph{Age} attribute is \emph{21}. Automatic, reliable information extraction is hard in such domains. More generally, illicit domains tend to frequently exhibit such heterogeneity; the findings in this paper would also apply to them.

A specific attribute that is extremely important to investigators is the \emph{geotag} implicitly conveyed by the webpage. Such tags are often present in free text fields like description or the page body, and not within structured HTML tags (hence, cannot be extracted by wrapper-based extractors \cite{wrapperIE}). Even though geolocations like cities are \emph{not} obfuscated\footnote{Advertisements want to be easily searchable on location facets, as trafficking victims change locations frequently.} in human trafficking pages, automatically extracting them is problematic both due to the language model, and due to \emph{ambiguity}. For example, \emph{Charlotte} may refer either to the city in North Carolina or to a person. Using a lexicon to directly extract geolocations is problematic for this reason; richer clues like context (such as the words surrounding an extraction) are necessary for disambiguation \cite{kejriwal2017information}, \cite{multitask}. Looking at the context of \emph{neWYOrk} in the earlier example, for instance, one can deduce that it most likely refers to a name, not the city.

We also note that, to infer a \emph{geotag} (referring to a single identifiable location in the world), \emph{prior knowledge} and \emph{relational information} can both prove necessary. For example, there is a \emph{Los Angeles} in both \emph{California} and \emph{Texas}. Given only a \emph{Los Angeles} geolocation extraction, one is more inclined to infer \emph{Los Angeles, California}, since it has much higher population than \emph{Los Angeles, Texas}. However, if \emph{Texas} were \emph{also} extracted as a geolocation, the probability of the latter increases due to the \emph{relational} connection. 

In this paper, we present a geotagging framework that holistically integrates the strengths of \emph{semantic lexicons}, \emph{extraction context}, \emph{relational constraints} and \emph{prior cues} like city populations to deliver high performance. The workflow is illustrated in Figure \ref{frameworkfig}. First, the corpus of pages is preprocessed using text scrapers and tokenization. Next, the Geonames dictionary is used to label tokens as geolocation \emph{candidates} \cite{geonames}. For example, Charlotte would be annotated as a geolocation candidate regardless of whether it is a name or a city in the underlying webpage. To determine the \emph{probability} of the candidate being a geolocation, we use a recent machine learning-based approach that uses context features \cite{kejriwal2017information}. The next few steps involve building and solving an Integer Linear Programming (ILP) model that integrates relational constraints and external domain knowledge (such as city populations) from Geonames as ILP constraints. After solving the model (per webpage), the result is a set of extractions that achieves high precision, without significantly hurting recall compared to non-ILP processing.
We apply the framework in Figure \ref{frameworkfig} to the problem of geotagging a corpus of real-world human trafficking webpages. 

{\bf Contributions.} We summarize our main contributions as follows. (1) We present a novel framework that integrates the strengths of several approaches to achieve good geotagging performance. (2) We show how domain constraints that are assumed as obvious by human beings can be encoded as constraints in an ILP model to improve performance over a machine learning-only approach. (3) We present some preliminary empirical results on a real-world human trafficking geotagging task illustrating the promise of the approach.    

{\bf Structure of the paper.} Section \ref{relatedwork} covers some related work, Section \ref{framework} describes the overall framework, including sub-components, Section \ref{experiments} describes the experimental results and Section \ref{futurework} concludes the work.   

\section{Related Work}\label{relatedwork}

The extraction of geolocation information such as cities and countries from unstructured data is an important problem that more generally falls within Information Extraction (IE). IE is an old research area for which a wide range of techniques have been proposed; for an accessible survey of Web IE approaches, we refer the reader to \cite{IEsurvey}.   

The goals of this work are similar to other geolocation prediction system for `difficult' datasets like Twitter, a good example being \cite{rev1ref1}. However, we note that illicit domain challenges are different from those of social media, an important example being \emph{information obfuscation} \cite{ben}, \cite{kejriwal2017information}, \cite{dig}.

The individual components that are used for building the framework are well established in the research community. For example, word embedding methods, used in the contextual classifier in our framework, have achieved notable advances in NLP (and especially IE) performance \cite{multitask}. Integer Linear Programming has also been used in many applications, and offers a powerful and flexible way to represent constrained optimization problems \cite{ilp}. Lexicon-based IE has also received much coverage in the literature, an influential recent work being \cite{lexiconIE}. Given its importance, geolocation extraction has received a lot of focused attention in the literature, an important related work being the recent text and context-based approach by Speriosu and Baldridge \cite{toponym}. Some of the techniques in this work, such as usage of text and populations, derive from extant techniques on \emph{toponym resolution} \cite{toponymbook}, \cite{toponym}. A good description may be found in the book by Leidner \cite{toponymbook}. We note, however, that except for a recent paper that we published \cite{kejriwal2017information}, no work has tackled the challenges of high-performance geolocation extraction in domains like human trafficking. This work improves the performance of our recent work by a significant margin by incorporating constraints into a geolocation-specific ILP model. Additionally, unlike our previous work, the system in this paper is optimized specifically for geolocation extraction, given its importance to human trafficking investigators.  

\section{Framework}\label{framework}
\begin{figure}
\centering
\includegraphics[height=2.6in, width=3.5in]{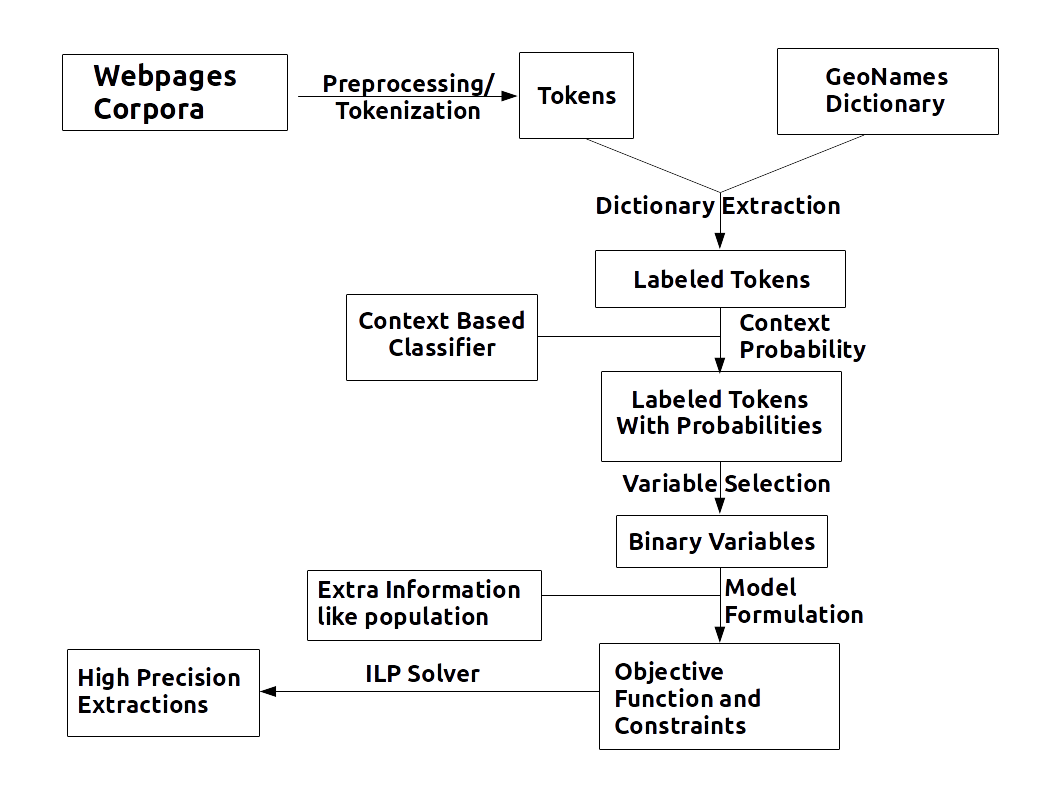}
\caption{A workflow-level illustration of the geotagging framework described in this paper}\label{frameworkfig}
\vskip -6pt
\end{figure}

The overall approach is illustrated in Figure \ref{frameworkfig}. The input to the system is a corpus of webpages, serialized as raw HTML obtained by a domain-discovery crawling system, and the final output is a set of high-precision geolocation extractions for each webpage. We detail the individual steps in the approach below. In this paper, we assume that the webpages are from the human trafficking domain, a difficult domain whose challenges were earlier described.

\subsection{Preprocessing}\label{preprocessing}
Since the relevant geotags in the webpage are typically present in \emph{natural language} elements like title, description and text (as opposed to structured fields), the first step is to extract the text from the webpage. This \emph{preprocessing} step is non-trivial, and involves automatically removing extraneous elements like HTML tags and irrelevant characters. Like other extractors, there is often a precision-recall tradeoff i.e. aggressively removing irrelevant information can also lead to the removal of relevant information. For that reason, we used an openly available \emph{tunable} text extractor called the Readability Text Extractor\footnote{\url{https://www.readability.com/}}, and optimized it for separately achieving high text extraction recall and precision. The \emph{lowercased} extracted text is \emph{tokenized} using whitespace and punctuation as delimiters. Extraction and tokenization are performed independently for webpage \emph{title}, using only the high precision setting, and \emph{main body text}, using both high recall and high precision settings.

As a running example, alluded to earlier in the introduction, the title and main text (i.e. \emph{body}) from the small HTML fragment \emph{<html> <head> <title> Los Angeles Escort Listing </title> </head><body> <p> My name is Charlotte </p>. I come from Mexico and am new in the city of Los Angeles </body> </html>} are separately preprocessed and passed to the next step as three lists of tokens (two for the text, and one for the title).

\subsection{Dictionary-based Candidate Extraction}\label{extraction}
Due to explosion of structured data on the Web, there are openly available \emph{dictionaries}, also called \emph{semantic lexicons} \cite{semanticlexicon}, for identifying \emph{candidate} geotags from text. For geotags, a standard near-comprehensive lexicon is Geonames \cite{geonames}. Because Geonames contains many geotags that are not interesting in our domain, such as unpopulated places, we limit our lexicon to a subset containing states\footnote{In Geonames, states are usually marked as \emph{first-level} administrative divisions.}, countries, and also those cities with a population greater than 15,000. 

For efficiency, we process the items in the lexicon using a trie data structure. Using the trie, we implement an exact string matching algorithm on the lists of tokens output by preprocessing. One reason for using the trie is that there are locations in the lexicon that span \emph{multiple} contiguous tokens (e.g., 'los angeles'); the trie can efficiently extract such \emph{token spans} as candidates. Using the running example, the token spans 'los angeles', 'charlotte', 'mexico', 'the city' and 'angeles' are all marked as city tokens\footnote{Both `mexico' and `angeles' are individually present in the \emph{City} subset of the Geonames lexicon.}, while 'mexico' is marked as a country token.

It is important to note that these candidate extractions are a (typically large) \emph{superset} of the true positive extractions. In that sense, the candidates are \emph{high-recall} (often, perfect-recall) and low-precision. There are multiple reasons for this phenomenon; we note three important ones. The first problem is the existence of generic or slang terms like `the city' in  the lexicon (present in Geonames because it is an alternative term for the city of London) that cause significant false positive extractions. The second problem is ambiguity e.g., \emph{Charlotte} is both a name and a city. The third, possibly most difficult problem, is that even when correctly extracted, a city might not be the relevant extraction. This last problem occurs when there are multiple cities that got extracted, but only one is relevant, usually the place where the subject of the webpage (in the human trafficking domain, an escort) is \emph{physically} advertising from.

\subsection{Context Based Classification}\label{classification}
To improve the precision of the candidates, we propose using the \emph{context} of the candidate in the text. Intuitively, even the \emph{local} context is often a very revealing clue to humans about whether, for example, `Charlotte' is a name or a city. In the running example, the words preceding Charlotte (`My name is') allow us to tag Charlotte as a name with near certainty. \emph{Context-based classification} marks candidate annotations as positive or negative using a supervised machine learning procedure that is trained on true positives and negatives using features derived from the context.     

In earlier work, CRFs, with manually crafted feature functions, were often used for this purpose. As described in our recent work, such feature functions are often problematic for irregular, obfuscated domains like human trafficking \cite{kejriwal2017information}. Another problem is the large number of annotations typically required for CRF-based taggers. In previous work, we presented a minimally supervised context-based classifier that avoids both the feature crafting and high supervision problems by first efficiently deriving low-dimensional \emph{word embeddings} from an extracted text corpora as \emph{word feature vectors} (WFVs) \cite{kejriwal2017information}. A contextual classifier is trained by using the WFVs to derive a contextual feature vector (CFV) for each candidate token span. Intuitively, this is done by aggregating and normalizing the vectors of tokens occurring in a window of 5 words (on either side of the candidate).

Using a small training set, a random forest classifier is trained to \emph{probabilistically} mark each candidate as correct. Because of the low dimensional feature space, even a small training set allows a classifier to quickly generalize.

In previous work, we used a threshold on the probability scores to determine which candidates were relevant. In Section \ref{experiments}, we consider a variant of this approach (choosing the most probable candidate from a webpage as the correct extraction) as a baseline. 

The hypothesis guiding the framework in Figure \ref{frameworkfig} is that systematically processing probability-annotated candidates using \emph{domain knowledge} can aggressively improve precision. Domain knowledge is captured as a set of \emph{constraints} used efficiently in an ILP framework to determine the subset of correct geotag candidates.

\subsection{Context-rich ILP Framework}
Before presenting the details of the approach, we introduce some formalism for ILP.

{\bf Definition (ILP).} Integer Linear Programming (ILP) is an optimization problem of the form:

\emph{Maximize:}
\begin{center}
$\sum\limits_{j=1}^n c_{j}x_{j}$
\end{center}

\emph{Subject to:}
\begin{center}
$\sum\limits_{j=1}^n a_{ij}x_{j} <= b_{i}$

$x_{j} \geq 0$

$x_{j} \in \mathbb{Z}$

\end{center}

Intuitively, ILP attempts to maximize an objective function subject to a given set of \emph{linear constraints} (in the formulation above, $i$ ranges over the set of constraints). A provably optimal solution to ILP is known to be NP-Complete; hence, solutions must be approximated. Good software packages for this problem already exist; we use a solution described further in Section \ref{experiments}. 

\subsubsection{Variable Selection}\label{variable}
At a high level, our framework encodes both candidate annotations and domain-specific constraints as ILP variables and constraints respectively. This involves some non-trivial modeling problems such as \emph{variable selection}.   

% The input for this step is the set of tokens from various parts of the web page (title and text in our case) along with city, state and country annotations and an assigned probability to each of the city tokens. We start the formulation of the problem in terms of an Integer Linear Programming solution.

We model the ILP variables $x_{j}$ in our framework as binary variables ($x_{j} \in \{0,1\}$). This reduces ILP to 0-1 linear programming, which is still NP-complete. Each variable is a placeholder for a candidate, with the simple semantics that a value of 1 (in an ILP solution) represents correctness with respect to its semantic type. In an ideal solution, for example, `Mexico' is correct with respect to semantic type \emph{Country} but not \emph{City}, for which it was also extracted by the lexicon as a candidate.

For consistency, we refer to the $x_j$ variables as \textit{token semantic type} (TST) variables. Note that, using the above example, if a token such as `Mexico' is marked with two semantic types, two TST variables will be created (mnemonically, \emph{Mexico-City} and \emph{Mexico-State}). However, if a token occurs multiple times in the extracted, preprocessed text, only one variable is created for the token. In other words, the \emph{number} of occurrences of a token is not taken into account in variable selection, as long as it occurs (i.e. marked as a candidate by the lexicon) at least once. 

{\bf Example:} The TST variables created for the running example in Section \ref{preprocessing} are:\emph{
Los Angeles - City,
Charlotte - City,
Mexico - City,
the city - City,
Angeles - City,
Mexico - Country.}

A problem with directly using TST variables as ILP variables is that, even if a particular city such as Los Angeles is set to 1, there could potentially be multiple cities in the world with the same name. Additionally, there is no way in the simple formulation to relate the city TST variables with the state and country TST variables. For \emph{canonical} geotagging, such \emph{relational} information is vital.

To accommodate this issue in our modeling, we introduce additional variables, denoted herein as \emph{composite TST} variables, to encode the intuitive notion that  a city is part of a state, and a state is part of a country. As before, we use Geonames for obtaining this relational information.

The new relational variables created are for each possible \emph{city-state}, \emph{city-country} and \emph{state-country} pair applicable for each candidate geotag. To clarify what we mean by applicable, we take the following scenario: suppose the respective sets of candidate states and cities annotated on a webpage are $S$ and $C$. To form applicable city-state composite TSTs, we take the cross-product of $S$ and $C$, and eliminate all pairs that do not occur in Geonames. In this way, we make novel use of Geonames as a \emph{relational lexicon} within our geotagging framework. 

Once created (by verifying against Geonames), the composite TST variables are included as additional ILP variables. For \emph{each} webpage, the set of (composite and non-composite) TSTs is the set of $x_j$ variables for an ILP model. Note that the relational information also leads to the introduction of new non-composite TSTs (at the state and country level) as the example below illustrated. Because we do not assume relational connections between webpages in this paper, each model is optimized independently, as described subsequently. 

{\bf Example:} For the running example, the composite TST variables are:\emph{
Los Angeles(city) in California(state),
Los Angeles(city) in Texas (state),
Charlotte(city) in North Carolina(state),
the city(city) in England(state),
Angeles(city) in Pampanga(state),
California(state) in United States(country),
Texas(state) in Unites States(country),
North Carolina(state) in United States (country),
England(state) in United Kingdom(country),
Pampanga (state) in Philippines(country).} 

The new non-composite TSTs introduced by the composite TSTs are \emph{California - State,
Texas - State,
North Carolina - State,
England - State,
Pampanga - State,
United States - Country,
United Kingdom - Country,
Philippines - Country.}

\subsubsection{Model Formulation}\label{ilp}

To fully specify the ILP model defined earlier, we need to formulate the objective function and the model constraints. This section describes both formulations. Note that the ILP model described below is independently constructed for each webpage, just like text preprocessing and candidate extractions. The model for each page can be optimized in parallel with other models, although we do not consider this option for the preliminary experiments in this work. 

{\bf Objective Function}. The weight ($c_j$ in the ILP definition) of variable $x_j$ in the objective function corresponds to how likely it is for the variable to be selected. The various factors that go into determining the weight are described below. We use two factors for the non-composite TST variables (\emph{token source} and \emph{context probability}), and two factors for the composite TST variables (\emph{population} and \emph{zero weight}).

\textit{Token Source.} The first factor is the part of the webpage that yielded the token list from which the candidate was extracted. From the experiments, we found a weight of 1.0 for title tokens lists, 0.5 for main body text extractor using \emph{strict} (i.e. high precision) setting and 0.4 for \emph{relaxed} main body text extractor to be optimal. Intuitively, if the candidate is extracted from the title, it is more likely to be the the correct city, and gets a higher weight.

\textit{Context Probability.} As described in Section \ref{classification}, each candidate has a probability associated with it, based on the output of the random forest classifier. Since a token might occur multiple times in the text, each with a different context and a different probability, we define the context probability weight of the candidate as the maximum over the context probabilities for all candidate occurrences within the page.

We take the average of the two weights above as the weight of the corresponding non-composite TST in the ILP model. The weight is guaranteed to be between 0.0 and 1.0. 
Next, we describe the factors determining composite TST weights:

\textit{Population.} Since multiple cities with the same name might occur in different states and countries, we bias our solution towards the city with the highest population. This is done by assigning population weights to city-state and city-country variables. For example the city Los Angeles occurs both in California and Texas in United States. Here, the variable Los Angeles - California would have a higher weight than the variable Los Angeles - Texas. The specific formula used to assign the weight $c\text{-}population_i$ for the composite TST variable $x_i$ is given by:

\begin{equation}
c\text{-}population_{i}= C_{pop} / K
\end{equation}

where K is a constant population factor (for normalization purposes) and $C_{pop}$ is the city population, which is obtained from Geonames.

\textit{Zero Weight.} Other composite variables are assigned a weight of 0, to make each selection equally likely. The likelihood that these variables are set to 1 by an ILP solver depends on the subsequently described model constraints.

{\bf Constraints.} We design a set of ILP constraints to limit candidate selection to extractions that are feasible and highly probable, using simple knowledge about cities, states and countries, and their relationships to each other.

\textit{Semantic Type Exclusivity.} A candidate in our framework can be one of a city, state, or country (i.e. a \emph{semantic type}) in an actual ILP model instantiation. Equationally, denoting $candidates$ and $types$ as the set of extracted candidates and relevant semantic types in our ontology,

\begin{equation}
\forall candidate_{i} \in candidates, \\
\sum\limits_{j=1}^{types} candidate_{i}type_{j} \leq 1 
\end{equation}

\textit{Number of Extractions of a Semantic Type} This constraint limits how many extractions should be selected for a particular semantic type. This number is 1 for our purposes, as trafficking ads very rarely have more than one geotag.

\begin{equation}
\forall type_{j} \in types,
\sum\limits_{i=1}^{candidates} candidate_{i}type_{j} \leq 1 
\end{equation}

\textit{City-State/Country Feasibility.} A final solution to the modeled ILP problem would only be feasible if a chosen city is actually \emph{present} in a chosen state or country\footnote{In the solution, the composite TST of such a `chosen' city-country or city-state pair would be set to 1.}. For example, if the variable for \emph{Los Angeles} is set to 1, and the variable for \emph{United States} is set to 0, the composite variable representing the pair \emph{Los Angeles-United States} should not be set to 1 in a meaningful solution. 

Formally, the constraint can be expressed by stating that for each country, the sum of all city-country variables must be less than the country variable; similarly for states. If a country variable is 0, no corresponding city-country variables can be set to 1. If it is 1, \emph{at most} one of the city-country variables can feasibly be set to 1 (because of objective maximization, \emph{exactly} one is set to 1).

\begin{equation}
\forall country_{j} \in countries, \\
\sum\limits_{i=1}^{cities} city_{i}country_{j} \leq country_{j} 
\end{equation}
\begin{equation}
\forall state_{j} \in states, \\
\sum\limits_{i=1}^{cities} city_{i}state_{j} \leq state_{j}
\end{equation}

\textit{City-State/Country Exclusivity.} This constraint ensures that the chosen city has exactly one corresponding city-state and city-country variable set to 1. For each city, the sum of the city-country variables is equal to the city variable. If the city variable is set to 1, exactly one of the city-country variables must be set to 1; if 0, none of the city-country variables can be 1 in a valid solution:

\begin{equation}
\forall city_{i} \in cities, \\
\sum\limits_{j=1}^{countries} city_{i}country_{j} = city_{i} 
\end{equation}
\begin{equation}
\forall city_{i} \in cities, \\
\sum\limits_{j=1}^{states} city_{i}state_{j} = city_{i} 
\end{equation}

\subsection{Extraction Selection}\label{selection}
With the model as specified earlier, we solve the ILP problem for each webpage. Because variables are binary in our model, each variable can be set to only 1 or 0. As earlier, we refer to the variables set to 1 as \emph{chosen} variables. 

For non-composite TST variables, the candidate underlying a chosen variable is marked as a correct extraction of its semantic type. Due to domain semantics, we permit at most one TST to be chosen per semantic type.

For composite TST variables, a chosen variable defines the underlying relationship to be correct. For example if \emph{Los Angeles(city) - in - California(state)} is set to 1, the selected candidate city is Los Angeles, the selected candidate state is California, and the relationship denotes that we are referring to Los Angeles in California. Because of exclusivity and feasibility constraints, note that all three variables must have been chosen for this to occur. Furthermore, the chosen relationships permit us to determine city geotags canonically, which is necessary both for visualization and for accurate geolocation analytics.
% \begin{table}
% \centering
% \caption{Stanford NER features that were used for re-training the model on our annotation sets}
% \begin{tabular}{|p{1.3in}|p{1.4in}|} \hline
% useClassFeature=true& useNext=true\\ \hline
% useWord=true& useSequences=true\\ \hline
% useNGrams=true& usePrevSequences=true\\ \hline
% noMidNGrams=true& maxLeft=1\\ \hline
% useDisjunctive=true&useTypeSeqs=true\\ \hline
% maxNGramLeng=6&useTypeSeqs2=true\\ \hline
% usePrev=true&useTypeySequences=true\\ \hline
% wordShape=chris2useLC & \\ \hline
% \end{tabular}\label{nerfeatures}
% \end{table} 

\section{Preliminary Experiments}\label{experiments}
\subsection{Setup}
{\bf Datasets.} The datasets for the experiments are sampled from webpages in the \emph{human trafficking} domain, crawled as a part of the DARPA MEMEX program\footnote{\url{http://www.darpa.mil/program/memex}}. 

Since we use the supervised contextual classifier described in our previous work \cite{kejriwal2017information}, in conjunction with the Geonames lexicon (described in Sections \ref{extraction} and \ref{classification}) \cite{geonames}, a \emph{training} dataset is required. We train the contextual classifier on a sample of 75 webpages with manually annotated geotags\footnote{Because the ILP is \emph{unsupervised}, these geotags are \emph{non-canonically} annotated; i.e. a geotag is simply marked as correct or incorrect, but the canonical geotag is unknown (at least directly) to the framework algorithm. In that sense, the training set is weaker than the test set, which is annotated with canonical tags.}.

We use 20 webpages for the test dataset, which are manually annotated with \emph{canonical} geotags (e.g., `Los Angeles, California, United States', which has a Geonames identifier).

{\bf Baselines.} We consider two baselines. We compare (using subsequently described metrics) the chosen candidate city per page to the correct city. The first baseline, \emph{Random}, randomly selects a candidate extraction as the correct one. The second baseline, \emph{Top Ranked}, chooses the candidate assigned the highest probability by the contextual classifier. Comparisons with \emph{Top Ranked} allows us to compute the effect of the ILP model on geotagging performance.

{\bf Metrics.} We report precision and recall on the test set to evaluate performance. 

{\bf Implementation.} We used the licensed version of Gurobi Optimizer \cite{gurobi} for modeling and solving ILP. All other code is written in Python 2.7. All experiments were run on a machine running 64bit Ubuntu 16.04 with Intel Core i7-4700MQ CPU @ 2.40GHz x 8 and 8GB RAM.

\subsection{Results}\label{results}
{\bf Comparison With Baselines} Table \ref{performance} shows the performance of the proposed framework against the two baselines. Unsurprisingly, the random baseline performs the worst, and using contextual classifiers provides a significant improvement in comparison. The additive effects of ILP to performance are promising, leading to improvements in both precision and recall.  

{\bf Error Analysis.} On further exploration of the wrong results, we found that ILP marks an incorrect candidate as correct when a city with high population is present in the footer (or other areas) of the webpage. We believe that, with better text extraction, this issue can potentially be mitigated (by recognizing a segment of the text as the footer, and ignoring it) or by assigning more weight to the main sections of the page.

Wrong candidates are also chosen when the name of a city in the lexicon corresponds to an \emph{alternate} name of a popular city. For example, in some cases, the candidate `the city' was labeled as the correct city. This is because the city of London is popularly known as `the city' by locals. The context in which the candidate `the city' is present also points to it being a city name e.g., `I have recently moved to the city'. Since the population of London is much higher than actual city the page is referring to, it is often selected as the correct geolocation of the page. This is a more difficult problem to avoid without ad-hoc engineering effort; we are currently investigating automated solutions to the problem. 

% {\bf Running Time} With the time taken for the solution to converge being one of the important concerns while using an Integer Linear Programming solution and the potential large amounts of datasets that the solution needs to be run on, the running time was an important metric to be noted. On the current dataset the solution converged in the mean duration of 0.364 secs.

\begin{table}
\centering
\caption{Performance comparison with baselines}
\begin{tabular}{|p{1.0in}|p{0.9in}|p{0.9in}|}\hline
{\bf Model}& {\bf Precision}& {\bf Recall}\\ \hline
Random& 0.5& 0.35714286\\ \hline
Top Ranked& 0.61538462& 0.57142857\\ \hline
ILP& 0.78571429& 0.78571429\\ \hline
\end{tabular}\label{performance}
\end{table}

\subsection{Discussion}\label{discussion}
Early results show that ILP provides a simple, unsupervised way to improve geotags output by upstream machine learning models. Although these results are preliminary, and experiments on more datasets and domains are required to validate them completely, we hypothesize that the main advantage of such a model is a systematic encoding of constraints that hold universally for the domain. In practice, the constraints are able to successfully deal with noisy candidates and candidate classifications by using exclusivity and feasibility constraints.  

Interestingly, the ILP model also allows us to encode other pieces of geographical information, such as city population, that are clearly important in the real world when identifying geolocations in the face of noisy and uncertain information. Other pieces of information, also readily available from Geonames and other knowledge bases, can also be included in a similar manner. Some of these were already used in the present system e.g., slang/local terms, like `the city' (for London), available in Geonames. An open issue is whether we can automatically distinguish between slang terms that lead to better recall without necessarily harming precision, and those that end up causing more noise (such as `the city'). A promising alternative, on which there is limited work, is to acquire more labels by crowd-sourcing to facilitate better training of the context-based classifier \cite{crowdsourcing}. We are currently investigating such possibilities in human trafficking geotagging. 

Finally, we note that, although the model was used primarily for geotagging, it can also be systematically extended to other semantic types such as names and nationalities. Early experiments on some other semantic types important for human trafficking have yielded promising results. We are continuing to investigate the model further.

\section{Future Work}\label{futurework}
We will integrate more constraints into the ILP based model to improve performance even further, as well as more experiments (using both more datasets and more illicit domains) to validate the early results in this paper. Active efforts are already underway to scale the system on many millions of scraped human trafficking webpages.  

{\bf Acknowledgements} This research is supported by the Defense Advanced Research Projects
Agency (DARPA) and the Air Force Research Laboratory (AFRL) under contract number FA8750-
14-C-0240. The views and conclusions contained herein are those of the authors and should not
be interpreted as necessarily representing the official policies or endorsements, either expressed
or implied, of DARPA, AFRL, or the U.S. Government.

%
% The following two commands are all you need in the
% initial runs of your .tex file to
% produce the bibliography for the citations in your paper.
\bibliographystyle{abbrv}
\bibliography{sigproc}  % sigproc.bib is the name of the Bibliography in this case
% You must have a proper ".bib" file
%  and remember to run:
% latex bibtex latex latex
% to resolve all references
%
% ACM needs 'a single self-contained file'!
%
%APPENDICES are optional
%\balancecolumns
%Appendix A

%\balancecolumns % GM June 2007
% That's all folks!
\end{document}